\documentclass[10pt,twocolumn,letterpaper]{article}

\usepackage{cvpr}
\usepackage{times}
\usepackage{epsfig}
\usepackage{graphicx}
\usepackage{amsmath}
\usepackage{amssymb}
\usepackage{booktabs}
\usepackage{caption} 
\usepackage{authblk}
\usepackage{fancyhdr}
\usepackage[
includehead,
nomarginpar,
textheight=22cm,
textwidth=17cm,
headheight=14.84526pt,
headsep=0.5cm,
]{geometry}
\usepackage[toc]{appendix}
\usepackage[pagebackref=true,breaklinks=true,letterpaper=true,colorlinks,bookmarks=false]{hyperref}

\cvprfinalcopy 

\fancypagestyle{titlepage}{
    \fancyhead[C]{\textbf{Preprint Notice:} This paper is a preprint of a paper submitted to the 26th Irish Machine Vision and Image Processing Conference (IMVIP 2024). The accepted copy of record will be available at IET Digital Library.}
    \fancyhead[L]{}
    \fancyhead[R]{}
    
    \fancyfoot[C]{\thepage}
    \fancyfoot[L]{}
    \fancyfoot[R]{}
}

\begin{document}

\renewcommand*{\Authsep}{\hspace{2em}}
\newcommand\CoAuthorMark{\footnotemark[\arabic{footnote}]} 
\author{Mohammad Abuzar Hashemi\thanks{equal contributions.}}
\author{Zhanghexuan Ji\protect\CoAuthorMark}
\author{Mihir Chauhan}
\author{Yan Shen}
\author{Abhishek Satbhai}
\author{Mir Basheer Ali}
\author{Dana Moukheiber}
\author{Sargur Srihari}
\author{Mingchen Gao}
\renewcommand*{\Authsep}{\authorcr}
\renewcommand*{\Authands}{\hspace{2em}}

\affil{Department of Computer Science and Engineering, University at Buffalo,\\
The State University of New York, Buffalo, NY, USA\\
{\tt\small \{zhanghex, mihirhem, yshen22, danamouk, srihari, mgao8\}@buffalo.edu} \\
{\tt\small {\{ma.hashemi.786, alimirbasheer, abhishek07satbhai\}}@gmail.com}
}

\title{LAViTeR: Learning Aligned Visual and Textual Representations\\ Assisted by Image and Caption Generation}
\maketitle
\thispagestyle{titlepage}  

\begin{abstract}

Pre-training visual and textual representations from large-scale image-text pairs is becoming a standard approach for many downstream vision-language tasks. The transformer-based models learn inter- and intra-modal attention through a list of self-supervised learning tasks. This paper proposes LAViTeR, a novel architecture for visual and textual representation learning. The main module, Visual Textual Alignment (VTA) will be assisted by two auxiliary tasks, GAN-based image synthesis and Image Captioning. We also propose a new evaluation metric measuring the similarity between the learnt visual and textual embedding. The experimental results on two public datasets, CUB and MS-COCO, demonstrate superior visual and textual representation alignment in the joint feature embedding space. Our code is available at \url{https://github.com/mshaikh2/MMRL}

\end{abstract}

\section{Introduction}

Learning cross-modal visual and textual representation is essential for bridging the semantic gap between images and languages. It is the cornerstone for a wide range of vision-language (V+L) tasks, such as image-text cross-modal retrieval, visual question answering (VQA) \cite{anderson2018bottom}, image captioning \cite{anderson2018bottom}, and so on.

\begin{figure}[ht]
\begin{center} 
  \includegraphics[width=1.0\linewidth]{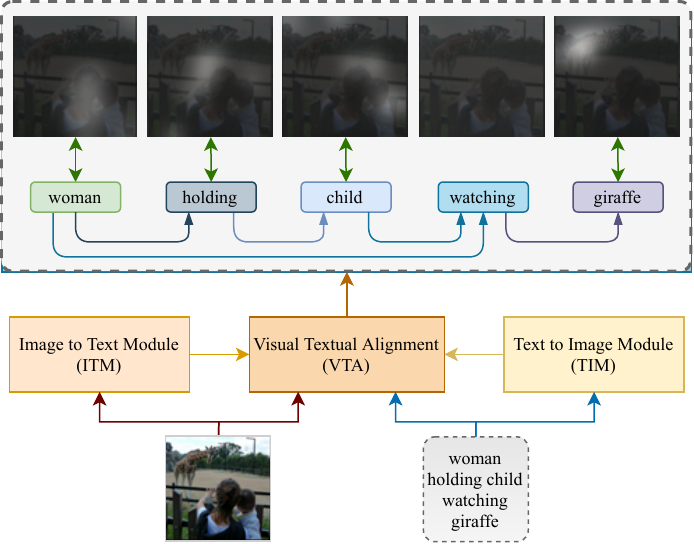}
\end{center}
\caption{An overview of the end-to-end LAViTeR network. VTA module is assisted by ITM and TIM modules, which in-turn learns to better align the corresponding visual and textual counterparts. The bidirectional arrows indicate the alignment between words and their respective objects in the given image. The intra-word arrows indicate the relationships between the input words that the network learns.}
\label{fig:overview_laviter}
\end{figure}

Inspired by the success of BERT \cite{bert} and XLNet \cite{yang2019xlnet} using self-supervised learning on natural language processing, there has been a surging research interest in vision-language pre-training on image-text pairs. The learned task-agnostic representation is shown to be effective for many image-language applications after fine-tuning on specific downstream tasks. Self-supervised learning is designed to explore the organization of the data as its own source of supervision. This promising approach releases the burden of annotating data with ground truth labels, provides an opportunity to explore a large amount of unlabeled data such as image-text pairs, video-text pairs in free form format from online platforms. This approach has been applied to radiology images combined with their associated reports \cite{li2020comparison,chauhan2020joint} to leverage the abundance of unlabeled medical data. This data can retrospectively be collected from clinical routine, and has a lot of potential for self-supervised learning.

The representation alignment can be roughly classified into two categories, one-to-one matching and many-to-many matching. One-to-one matching focuses on the global representation from images and sentences, and then associates them by exploiting visual-semantic embedding \cite{wang2018learning}. Many-to-many matching methods incorporate relationship between regions of a image and words of a sentence to capture fine-grained cross-modal matching \cite{huang2017instance}. 

Analogous to the pre-training task in BERT, some pre-training tasks for image-text pairs include the Masked Language Modeling conditioned on image regions and Masked Region Modeling conditioned on input text. Those approaches randomly mask some words or regions from the input and use a Transformer model to recover the words or regions. Many of the fine-grained region-word matching rely on the modern object detectors \cite{chen2020uniter,li2020oscar,wei2020multi}, usually Faster R-CNN \cite{ren2015faster}, to detect salient regions and match them to words. However, the state-of-the-art object detection, which needs to leverage large amount of annotated bounding boxes for supervised learning, is not always available for domain-specific datasets. 

Motivated by the above discussion, we propose a model for \textbf{L}earning \textbf{A}ligned \textbf{Vi}sual and \textbf{Te}xtual \textbf{R}epresentation (\textbf{LAViTeR}). As shown in Figure \ref{fig:overview_laviter}, the main goal of LAViTeR is to learn the joint multi-modal embedding using visual textual alignment (VTA) module, which is assisted by two other self-supervised modules, the text-to-image module (TIM) and image-to-text module (ITM). The method was inspired by CycleGAN \cite{zhu2017unpaired} and its extension MirrorGAN \cite{qiao2019mirrorgan} on image synthesis. In our model, the given images and text are encoded to generate corresponding text and images, respectively. The generated text and images are trained to be mapped back to the original images and text in a cycle. Not only the features learned from real image-text pairs are aligned in the VTA module, the features learned from real-image-fake-text pairs and the fake-image-real-text pairs will also assist the representation learning and alignment. These generated images and texts can provide much more samples outside the training set and make our model more diverse and robust to changes in real data. The proposed method uses high-level features without any explicit supervision, avoiding explicit object detection. Our approach is suitable for the situations where the state-of-the-art object detection model is not feasible, and where bounding box annotations are not available for training.

Our contributions are summarized as follows: 1) We introduce LAViTeR for the image-text representation for V+L tasks. 2) We introduce two auxiliary pre-training tasks, GAN-based image synthesis and image captioning, to assist the representation learning. 3) We propose to use a new metric to quantitatively evaluate the similarity between the image and text representation in the embedded space.

\section{Related Work}
In this section, we discuss related work about multi-model representation learning and alignment. We also briefly review two main tasks in our architecture to pre-train the joint representation, the GAN-based image generation and image captioning. 

\subsection{Multi Model Representation Learning}
VilBERT \cite{lu2019vilbert} and LXMERT \cite{lxmert} are the two pioneering works in image-text joint representation learning, utilizing two streams of Transformers to images and text independently. Those image and text representations then fused by a self-attention mechanism in the later stage. After those two pioneered work, single-stream architecture has also been proposed using a single Transformer to jointly learn image-text embedding, such as UNITER \cite{chen2020uniter}, VisualBERT \cite{li2019visualbert}, Unicoder-VL \cite{Unicoder}, VL-VERT \cite{su2019vl}, B2T2 \cite{alberti2019fusion}. Typical self-supervised learning tasks, such as masked language modeling, masked region modeling, image-text matching, and word-region alignment are applied to pre-train the models. More recently, VILLA has been proposed using adversarial training  as a general framework can be applied to any V+L models \cite{gan2020large}.

There are quite a lot of work trying to explicitly enforce the word and region alignment, such as VisualBERT \cite{li2019visualbert}, UNITER \cite{chen2020uniter}, Oscar \cite{li2020oscar}, MMAC \cite{wei2020multi}. For example, UNITER \cite{chen2020uniter} uses the Optimal Transport \cite{peyre2019computational} to explicitly calculate the minimum cost of transporting between the image embedding to word embedding.

\subsection{Text to Image Generation}
Image synthesis from text is a fundamental task in multi-modal learning across vision and language. Most proposed works in image generation are based on conditional Generative Adversarial Networks (GAN). A common approach utilizes a text encoder, mostly RNN-based text encoder before the popularity of Transformer, to encode the text description to guide the image generator \cite{reed2016generative, stackGan}. The attention mechanism is widely used to guide the generator to focus on certain words when generating specific regions \cite{xu_attngan}. The attention mechanism is used to capture the similarity between the generated images and the sentences in both the global level and fine-grained word/region level. MirrorGAN \cite{qiao2019mirrorgan}, tries to learn better text-to-image generation by re-describing the generated images.


\subsection{Image Captioning}
Image Captioning is the reverse process of text-to-image generation. It typically consists of a CNN encoder and an RNN decoder to transfer the information from images to the generated text description \cite{vinyals2015show}. Attention mechanism has been shown very effective focusing on salient objects while generating the corresponding words \cite{xu2015show}. Following the success of Faster-RCNN \cite{ren2015faster} in object detection, bottom-up features provide informative regions in the image, which are used for region level attention \cite{anderson2018bottom, aoa2019} and visual scene graph modeling \cite{yao2018exploring}. Recently, transformer architecture \cite{vaswani2017attention} is also used for image captioning, which further boosts the captioning performance with its implicit self-attention mechanism \cite{caption2020imagetransformer, caption2019obj2word, ZhouPZHCG20}.




\begin{figure*}
\begin{center}
  \includegraphics[width=0.99\linewidth]{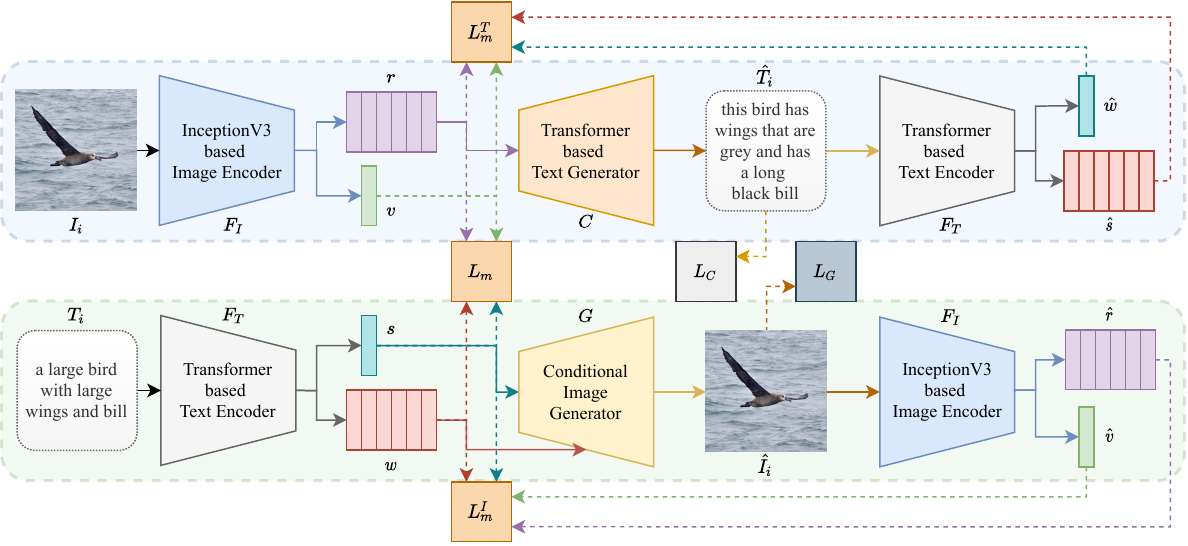}
\end{center}
   \caption{The architecture of the proposed LAViTeR. The pipelines with dotted outlines are the two assisting tasks, namely image to text and text to image conversion. Feature vectors of real image regions are indicated by $r$ while $v$ denotes the global image feature vector. Real text sentence level feature vector is indicated by $s$ while $w$ denotes the word level feature vectors. Similarly all $\hat{w},\hat{s},\hat{r},\hat{v}$ indicates the features extracted from generated samples. $L$ stands for various losses. Dotted arrows indicate the vectors that contribute the loss. Solid arrows indicate the vectors are input to the subsequent network.}
\label{fig:laviter_arch}
\end{figure*}

\section{Learning Aligned Visual and Textual Representations}
As shown in Figure \ref{fig:laviter_arch}, our proposed Learning Aligned Visual and Textual Representations (LAViTeR) model consists of three modules: the visual text alignment (VTA) module, which contains image encoder $F_I$ and text encoder $F_T$; the text-to-image (TIM) module, which is the image generator $G$; and image-to-text (ITM) module with the text generator $C$. VTA learns visual-text representations via matching real image-text pairs as our main task. In addition, we introduce two novel assisting tasks, illustrated in blue and green dashed boxes, for a better representation learning: fake-image-real-text matching with fake images generated from TIM; and real-image-fake-text matching where fake texts are converted from ITM using VTA image representations. Meanwhile, both TIM and ITM are also trained on their own losses. We will introduce each of them in the following part of this section. 

\subsection{VTA: Visual Text Alignment}
\label{sub:vta}
Given an image-text tuple $\langle I_i, T_i \rangle$, we want to learn the alignment between the words in $T_i$ and the parts of image $I_i$. 
For this, we first extract the global features $v$ and local features $r$ using an InceptionV3 \cite{cnn_inception} based image encoder, where $r \in \mathbb{R}^{D \times M}$ is flattened from the intermediate feature map of `mixed\_6e' layer and $v \in \mathbb{R}^{D}$ from the last average pooling layer. Both of them are projected to the representation feature space with a projection layer. This is denoted as a function $F_I$ in Figure \ref{fig:laviter_arch}, such that $r, v = F_{I}(I)$. Next, we extract sentence and word level features $s$ and $w$ respectively using a Transformer \cite{vaswani2017attention} based text encoder. In Figure \ref{fig:laviter_arch}, this text encoder is denoted as a function $F_T$, such that, $w, s = F_{T}(T)$, where $w \in \mathbb{R}^{D \times N}$ and $s \in \mathbb{R}^{D}$.

We define the $F_T$ as follows: Given text $T\in \mathbb{R}^{N}$, we use word-based token to embed it as $e\in \mathbb{R}^{D\times N}$, which is summed with positional encoding as the input of $F_T$. In transformer encoder layer, $e$ is first transformed into queries $Q=W^T_Qe$, keys $K=W^T_Ke$ and values $V=W^T_Ve$ within each attention head, where $W_Q, W_K, Q_V\in \mathbb{R}^{D\times D_k}$ in our setting. Then ``scaled dot-product self-attention'' is applied to $e$ as follows:
\begin{equation}
    \text{Attention}(Q,K,V)=f_s(\frac{QK^T}{\sqrt{D_{k}}})V
    \label{eq:attenc_head}
\end{equation}
where $f_s$ is the softmax function. 

Multi-head attention is applied to the self-attention sub-layer and the outputs from $h$ heads are concatenated: 
\begin{equation}
\begin{aligned}
    \text{head}_i = \text{Attention}(W^T_{Q_i}e,W^T_{K_i}e,W^T_{V_i}e)\\
    \text{Multihead}(e) = W_O\text{Concat}(\text{head}_1,...,\text{head}_h)
\end{aligned}
\label{eq:attenc_multihead}
\end{equation}
where $W_O\in \mathbb{R}^{D\times hD_k}$. 

The output from the multi-head attention is then sent to a feed-forward network. The residual mechanism \cite{cnn_resnet} and layer normalization are applied to multi-head attention and FFN outputs: 
\begin{equation}
\begin{aligned}
    \Tilde{e} &= \text{LayerNorm}(e+\text{Multihead}(e)) \\
    w &= \text{LayerNorm}(\Tilde{e}+\text{FFN}(\Tilde{e}))
\end{aligned}
\label{eq:attenc_output}
\end{equation}
and $s = \bar{w}$ is used as the sentence representation feature. 

Similar to \cite{xu_attngan} we align the word to image regions by implementing a word-level attention mechanism. First, the word-region  attention score $\alpha$ is obtained by multiplying the query, $w$ with context, $r$, and then normalizing the product using softmax. Next $\alpha$ is multiplied with the context $r$ to obtain the contextual vector $c$.
\begin{equation}
\begin{aligned}
m &= w^T \cdot r \\
\alpha &=  f_{s_{M}}(\gamma_1f_{s_{N}}(m)) \\
c &= r \odot \alpha
\end{aligned}
\label{eq:context_aware_vector}
\end{equation} 
Where $m \in \mathbb{R}^{N \times M}$ is the match vector; $f_{s_{N}}$ is the soft-max operation along $N$ words in text $T_i$; $f_{s_{M}}$ is the soft-max operation along $M$ sub-regions of image $I_i$; $\gamma_1$ is a hyper-parameter to tune the required amount of visual attention for a word and $\odot$ is a matrix multiplication operation.
Next, we calculate the element wise cosine similarity between $c$ and $w$ as $cos = (c^T w)/(||c||||w||)$ and compute the image to text matching score $S$ by following the work done in \cite{minClassErrorSpeech,discriminative}:
\begin{equation}
\begin{aligned}
S(I_i,T_i) = \log(\sum_{i=1}^{N-1}exp(\gamma_{2}cos))^{\frac{1}{\gamma_2}}
\end{aligned}
\label{eq:loss_image_sentence}
\end{equation}
where $\gamma_2$ is the importance magnification factor of the most relevant word and image sub-region in the given pair $\langle I_i, T_i\rangle$.

Finally, inspired by to \cite{fang2015captions,huangSemantic}, we calculate the the posterior probability $P$ of image $I_i$ matching with text $T_i$ in a batch of $B$ paired samples:
\begin{equation}
\begin{aligned}
P(I_i,T_i) &= f_{s_{B}}(\gamma_3S(I_i,T_i)) \\
L_{m_{w}}^{IT} &= -log(P(I_i,T_i))
\end{aligned}
\label{eq:prob_word_region}
\end{equation}
where $f_{s_{B}}$ is the soft-max of matching score $S$ over $B$ paired samples and $\gamma_3$ is a hyper-parameter and $L_{m_{w}}^{IT}$ is the loss when features of image sub-regions $r$ are matched to features of words $w$ in text. Here, the text and image samples at different index are considered as negative pairs and the samples at same index are considered positive pairs.
To maintain the symmetry, we also calculate $L_{m_{w}}^{TI}$ where the image and text variables are switched:
\begin{equation}
\begin{aligned}
P(T_i,L_i) &= f_{s_{B}}(\gamma_3S(T_i,L_i)) \\
L_{m_{w}}^{TI} &= -log(P(T_i,L_i))
\end{aligned}
\label{eq:loss_symmetry}
\end{equation}
Furthermore, we calculate the sentence level matching loss $L_{m_{s}}^{IT}$ by computing cosine similarity between global vectors $v$ and $s$ as $cos = (v^T s)/(||v||||s||)$ and substituting the value of $cos$ in Equation \ref{eq:loss_image_sentence}, \ref{eq:prob_word_region} and \ref{eq:loss_match}. Similar to before we can calculate $L_{m_{s}}^{TI}$ by switching $s$ and $v$. Thus we compute the total matching loss by adding all the losses.
\begin{equation}
\begin{aligned}
L_m = L_{m_{s}}^{TI} + L_{m_{s}}^{IT} + L_{m_{w}}^{TI} + L_{m_{w}}^{IT}
\end{aligned}
\label{eq:loss_match}
\end{equation}
During the first phase of training the objective is to reduce $L_{m}$ for pairs of real image $I$ and corresponding text $T$. For this training we preset $\gamma_1,\gamma_2,\gamma3$ as per the settings defined in \cite{xu_attngan} and the batch size $B=8$.

\subsection{TIM: Text to Image Module}
\label{sub:tim}
To learn the parameters for transforming the textual domain to visual counterpart we use Conditional Generative Adversarial Networks  \cite{gan, mirza2014conditional} with the sentence vector $s$ as the conditional input. Inspired by the AttnGAN \cite{xu_attngan}, we employ a cascade of GANs. Formally, our TIM task has $k$ discriminators with coupled generators $G \in \{G_1 \dots G_k\}$ that generate images $\hat{I} \in \{I_{i1} \dots I_{ik}\}$ of different scales, where the suffix $i$ indicate the $i^{th}$ datapoint. We utilize only the output of the $k^{th}$ generator for calculating the assisted losses explained in Equation \ref{eq:assist_loss}.

We first transform a real text sample $T_i$ using the transformer \cite{vaswani2017attention} based text encoder $F_T$ to output a sentence vector $s$ and word vector $w \in \mathbb{R}^{D \times N}$. We denote the entire image generation system, including generators and discriminators as a function $G$ as displayed in Figure \ref{fig:laviter_arch}. A 1-D uniformly sampled random noise vector $z$ along with $w$ as condition, are merged and fed as input to $G$ which outputs an RGB image $\hat{I_i}$ corresponding to $T_i$

The image generative loss $L_{G}$ is thus calculated in an adversarial setup and is defined as follows:
\begin{equation}
\resizebox{0.85\hsize}{!}{
$L_{G}= 
 - \frac{1}{2}\mathbb{E}_{\hat{I} \sim P_{G}}[\log(D(\hat{I}))]
- \frac{1}{2}\mathbb{E}_{\hat{I} \sim P_{G}}[\log(D(\hat{I},s))]$
}
\label{eq:loss_image_gen}
\end{equation} 
We train the discriminators, to learn to distinguish between the real $I$ and fake $\hat{I}$ samples, alternately with the generators while reducing the cross-entropy loss as below:
\begin{equation}
\resizebox{0.9\hsize}{!}{%
$
\begin{aligned}
L_{D} =
-\frac{1}{2}\mathbb{E}_{I \sim P_{data}}[ \log(D(I))] -\frac{1}{2}\mathbb{E}_{\hat{I} \sim P_{G}}[ \log(1-D(\hat{I}))] \\
-\frac{1}{2}\mathbb{E}_{I \sim P_{data}}[ \log(D(I,s))] -\frac{1}{2}\mathbb{E}_{\hat{I} \sim P_{G}}[ \log(1-D(\hat{I},s))]
\end{aligned}
$%
}
\label{eq:loss_image_disc}
\end{equation}
where $I$ is from the read data distribution $P_{data}$ and $\hat{I}$ is from the generated data distribution $P_{G}$. In Equation \ref{eq:loss_image_disc} the first and the second terms are unconditional and conditional losses as defined in \cite{stackGan}. Furthermore, we calculate the matching loss $L^I_m$ between fake image vectors $\hat{r}$, $\hat{v}$ and real text vectors $w$, $s$ respectively, similar to the process outlined in Subsection \ref{sub:vta}. Hence the total loss for the TIM can be defined as:
\begin{equation}
L_{TIM} = L_{G} + \lambda_{\hat{I}} L_{m}^I
\label{eq:loss_multitask}
\end{equation}
where $\lambda_{\hat{I}}$ is the hyper-parameter which is tuned to get better performance and the matching loss $L_m^I$ between the generated image $\hat{I}_i$ to real text $T_i$, as explained in Subsection \ref{sub:assisting_losses}.

\subsection{ITM: Image to Text Module}
\label{sub:itm}
The Image-to-Text branch (ITM) aims to generate fake text to assist VTA training in \ref{sub:vta}. It has been shown that the attention mechanism leverages the performance of image captioning models \cite{xu2015show}. Inspired by the power and success of the transformer layer in various vision tasks \cite{aoa2019, caption2019obj2word, caption2020imagetransformer, detr}, we decide to use a transformer based image captioning model $C$ as ITM. 

Following the implementation of \cite{vaswani2017attention}, we use a stack of transformer layers for the transformer encoder $C_e$ and decoder $C_d$. Similar to \cite{xu2015show}, in order to pass image regional features into the transformer, we use the flattened regional feature sequence $r\in\mathbb{R}^{D\times M}$ from $F_I$ as the input to $C_e$, where each column of $r$ is a representation corresponding to a certain part of the image. As for the decoder, given the caption $T_{1:N}$ of length $N$, all the words prior to the target position $p$ are embedded as $e_{1:p-1}\in \mathbb{R}^{D\times (p-1)}$, which is used as the decoder inputs to predict the $p$th word $\hat{T}_p$. 

According to \cite{detr}, when passing the image features into the transformer, in order to supplement the permutation-invariant issue, it's better to add positional encodings to the input of each self-attention sub-layer in $C_e$ and encoder-decoder cross-attention sub-layer in $C_d$ instead of only applied at the bottom of encoder. Similarly, the positional embeddings for text input are also added to the input of each self-attention sub-layer in $C_d$. 

Taking $r$ as input, $C_e$ applies the same operation as Eq. (\ref{eq:attenc_head}-\ref{eq:attenc_output}) in each transformer layer and refines the regional visual features via self-attention mechanism. Then it's top output $r_e=C_e(r)$ is passed to the encoder-decoder cross-attention sub-layer within each transformer layer of $C_d$ to further introduce the visual-language cross-attention mechanism for image-to-text generation. The sub-layer takes $r_e$ as keys along with the self-attention sub-layer output $\Tilde{e}_{1:p-1}$ as queries:
\begin{equation}
\begin{aligned}
    \Tilde{e}'_{1:p-1} &= \text{Multihead}(W^T_Q\Tilde{e}_{1:p-1}, W^T_Kr_e, W^T_Vr_e)
\end{aligned}
\label{eq:attdec}
\end{equation}
The decoder output is sent to MLP to predict the probability of word at position $p$:
\begin{equation}
    p(\hat{T}_p|r,T_{1:p-1})=f_s(\text{MLP}(C_d(C_e(r),e_{1:p-1})))
    \label{pred_txt}
\end{equation}

Given the ground truth caption $T_{1:N}$, we train the ITM with cross-entropy loss:
\begin{equation}
    L_{ITM}=L_C = -\sum_{p=1}^{N}log(p(T_p|r,T_{1:p-1}))
    \label{cap_loss}
\end{equation}

\subsection{Assisting losses}
\label{sub:assisting_losses}

Given the generated images $\hat{I}$ from TIM in Subsection \ref{sub:tim} and generated texts $\hat{T}$ from ITM in Subsection \ref{sub:itm}, we introduce two assisting matching losses in the model training: fake-image-real-text matching loss $L_m^I$ and fake-text-real-image matching loss $L_m^T$. 

Similar to $L_m$ in Subsection \ref{sub:vta}, we input $\hat{I}$ to $F_I$ and $\hat{T}$ to $F_T$ to get $\hat{r}, \hat{v} = F_I(\hat{I})$ and $\hat{w}, \hat{s} = F_T(\hat{T})$. From Eq. (\ref{eq:context_aware_vector}-\ref{eq:loss_match}), we can calculate the symmetric matching posterior probabilities between $(\hat{I},T)$ and $(I,\hat{T})$, thus $L_m^I$ and $L_m^T$ are as:
\begin{equation}
\begin{aligned}
    L_m^I &= L_{m_{s}}^{T\hat{I}} + L_{m_{s}}^{\hat{I}T} + L_{m_{w}}^{T\hat{I}} + L_{m_{w}}^{\hat{I}T}\\
    L_m^T &= L_{m_{s}}^{\hat{T}I} + L_{m_{s}}^{I\hat{T}} + L_{m_{w}}^{\hat{T}I} + L_{m_{w}}^{I\hat{T}}
\end{aligned}
\label{eq:assist_loss}
\end{equation}

Along with $L_G$ and $L_C$, our final multimodal loss for joint training is:
\begin{equation}
    L = \lambda_m L_m + \lambda_{\hat{I}} L_m^I + \lambda_{\hat{T}} L_m^T + \lambda_G L_G + \lambda_C L_C
    \label{eq:multimodal_loss}
\end{equation}
where $\lambda_m, \lambda_{\hat{I}}, \lambda_{\hat{T}}, \lambda_G, \lambda_C$ are hyper-parameters to add weights for each loss above.

\section{Experimental Results}
In this section, we explain our experiment settings, evaluation metrics and results to evaluate our proposed model. 

\subsection{Datasets} 
Our model is evaluated on two public datasets including CUB \cite{cub2011} and MS-COCO \cite{coco2014}, which are widely used in text-to-image generation and image captioning tasks. 

\textbf{CUB} is a dataset contains 200 bird species and is popular for classification, text-based image generation and image captioning tasks. It has 8855 training images and 2933 test images, where each image has 10 text descriptions. Since CUB only contains bird images with various attributes, the semantic domain is relative simple and narrow for image text representation, which is suitable for evaluating our model in a specific semantic task. We preprocess the dataset according to the method in \cite{xu_attngan}.

\textbf{MS-COCO} is a challenging dataset which is popular for various image-text related tasks including image captioning and image-text matching. It has 82783 training images which paired with 5 captions per image and 40504 images for testing. It provides large amounts of common object classes in the images, which can show the representation performance of our model in generalized semantic space.

\subsection{Evaluation Metric}
\label{sec: metric}
We employ R-precision a technique proposed in \cite{xu_attngan} to rank the retrieval results. To evaluate the task of reducing the heterogeneity gap between relevant word and image representation, we also propose an Attribute to Image Matching Cosine Score (AIMCoS) which matches the similarity of specific textual attributes which are supposed to be found in the paired image. Specifically, first for CUB dataset, we extract the attributes corresponding to each image file from the validation dataset. Each attribute contains text entries that list the features like color, bill length, shape, etc. of a bird. For each experiment, we first extract the global features $v$ of the image $I_i$ using $F_I$. Next, we extract the features $s$ of textual input of each attribute associated with image. Afterwards,  we compute the average of cosine similarity $f_{cos}$ of $v$ with the representations of each attribute corresponding to $I_i$.  Finally, we find the mean of match scores for all the images in the validation set. Mathematically if there are $K$ attributes present in image $I_i$ then
\begin{equation}
    AIMCoS = \frac{1}{U}\sum_{i=0}^{i=U}\frac{1}{K}\sum_{k=0}^{k=K}f_{cos}(v,s_k)
    \label{eq:aimcos}
\end{equation}
where $U$ is the number of images in the validation set of CUB dataset.

For computing the AIMCoS using $F_I$ and $F_T$ trained on COCO dataset, each of the 80 classes are considered as attributes. We then create a smaller evaluation set (LAViTeR\textsubscript{cocoeval}), which contains 100 images,  extracted from ImageNet \cite{deng2009imagenet} dataset queried on the keyword. Next, similar to Eq. \ref{eq:aimcos} we calculate the mean of $f_{cos}$ between the representations of each 100 images and the representation of text the respective class name. Lastly, this score is averaged over all the classes to calculate AIMCoS for COCO dataset. 
\begin{figure}[ht]
\begin{center} 
  \includegraphics[width=1.0\linewidth]{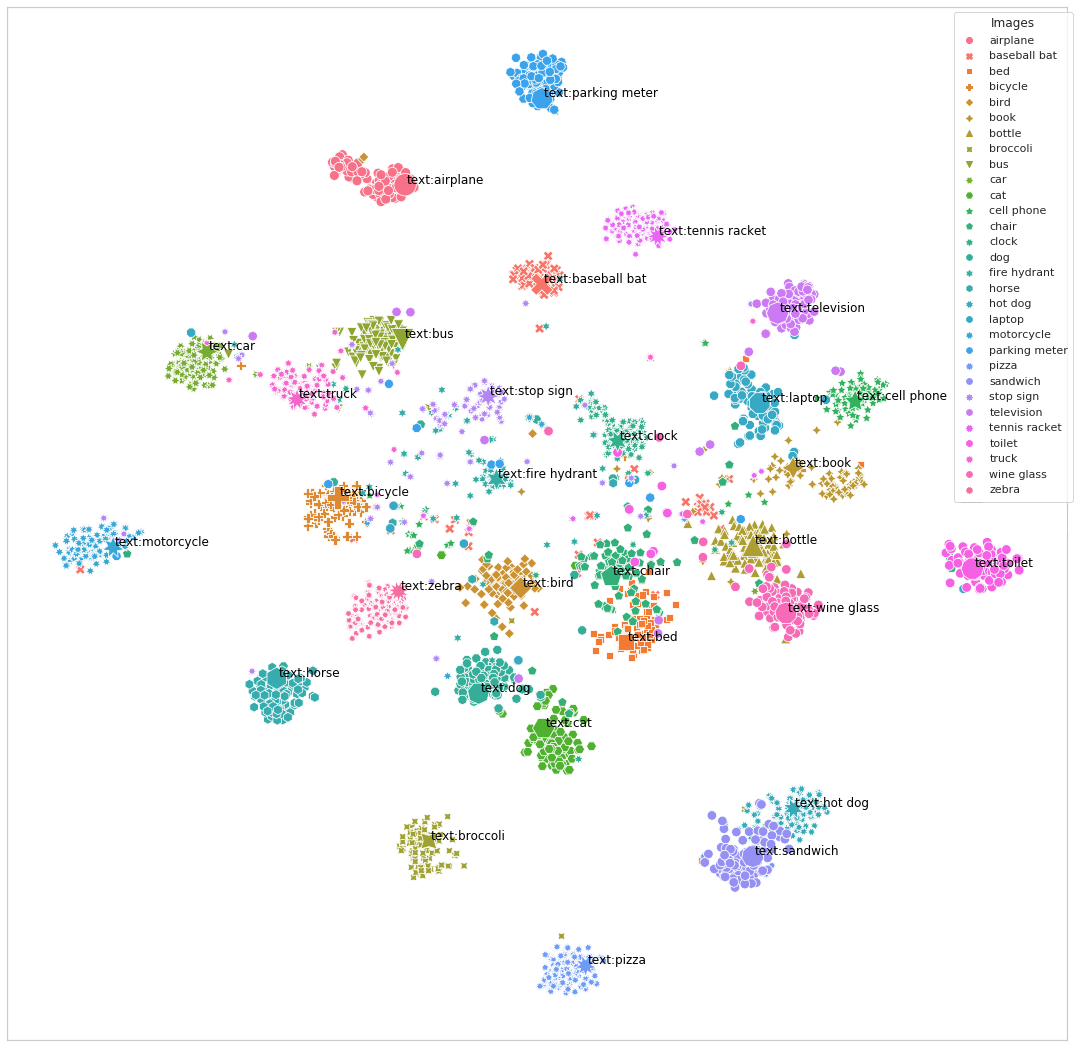}
\end{center}
\caption{A T-SNE \cite{maaten2008tsne} visualization of 3200 image  representations and 32 textual-label representations from LAViTeR\textsubscript{cocoeval}.}
\label{fig:tsne}
\end{figure}
In Figure \ref{fig:tsne}, the larger point with label present within each cluster is the representations of the textual labels. The other points of the cluster are the representations of image data-point. The LAViTeR model is able to bridge the representational gap between textual and visual data effectively while still performing at par with other techniques.

\subsection{Implementation Details}
The LAViTeR model is implemented in Pytorch \cite{torch2019} and all the experiments are carried out on NVIDIA GTX 1080ti GPUs. For $F_I$, we use the InceptionV3 model pretrained on ImageNet \cite{deng2009imagenet} as our backbone initialization. We use one transformer layer with 8 heads in $F_T$ and 6 transformer layers with 8 heads in both $C_e$ and $C_d$. The length of $r$ is 289 (flattened from $17\times 17$ feature map in $F_I$) and 15 for sentence length. The dimension of the word embedding and representation features $r, v, w, s$ from VTA is 256. Three cascaded generators are used in TIM and generate fake images with resolutions $64\times64$, $128\times128$ and $256\times256$ progressively. 

Three training phases are designed for LAViTeR in order to make the joint training more stable and easier to converge. In phase 1, we pretrain VTA on the training data so that $F_I$ and $F_T$ can generate acceptable image and text representations for the following phases. We freeze $F_I$ except the output layers for $r,v$, and train the transformer layer in $F_T$ and the output layers in $F_I$ with a learning rate 0.0002. As for phase 2, $C$ in ITM and $G$ in TIM are pretrained separately with the $r,v,w,s$ features from the pretrained VTA as inputs. The learning rate is set to 0.0002 for $G$, and for pretraining $C$ it is first set to 0.0001 and then decay by 0.1 after 20 epochs. When the pretrainings are completed, we freeze the first 5 layers of $F_I$ and jointly train all the modules together as our final LAViTeR model in phase 3, with a small learning rate of $10^{-6}$. All the training are optimized with the Adam optimizer \cite{adam2014} with a weight decay $0.0001$. The batch size is set to 96 for phase 1, 32 and 14 for $C$ and $G$ pretraining in phase 2, and 8 for the joint training due to the limitation of our GPU memory. In all experiments, $\gamma_1=4, \gamma_2=5, \gamma_3=10$ are used for the matching losses. 

\subsection{Performance}
\textbf{Hyper-parameter Selection} We first adjust the $\lambda$ in our multi-model loss (see Eq. \ref{eq:multimodal_loss}) in order to test the effect of different loss terms in our joint training and select the best $\lambda$ setting for our multi-modal loss. The results on CUB and COCO are shown in Table \ref{tab: hyperparam}. 

\begin{table}[!htp]\centering
\small
\begin{tabular}{lllllll}\toprule
\textbf{$\lambda_G$} &\textbf{$\lambda_C$} &\textbf{$\lambda_{m_{T}}$} &\textbf{$\lambda_{m_{I}}$} &\textbf{$\lambda_m$} &\textbf{R-precision} &\textbf{AIMCoS} \\\midrule
0 &0 &0 &0 &1 &89.76 &0.5045\\
1 &1 &1 &1 &50 &90.35 &0.4931 \\
0.01 &0.1 &1 &1 &50 &90.71 &0.5063\\
0.01 &0.1 &50 &1 &1 &90.31 &0.5063\\
0.01 &0.1 &1 &50 &1 &90.49 &0.496\\
0.01 &0.1 &10 &1 &10 &90.05 &0.498\\
0.01 &0.1 &10 &1 &1 &90.19 &0.501\\
0.001 &0.1 &1 &0.1 &10 &90.16 &0.501\\\midrule
\textbf{0.01} &\textbf{0.1} &\textbf{1} &\textbf{1} &\textbf{10} &\textbf{90.79} &\textbf{0.504} \\\midrule
1 &1 &1 &1 &1 &63.4 &0.203 \\
0.01 &0.1 &1 &5 &1 &62.2 &0.198\\
0.01 &0.1 &5 &1 &1 &64.6 &0.197\\
0.01 &0.1 &1 &1 &1 &63.0 &0.2\\
0.01 &0.1 &1 &1 &10 &66.4 &0.202\\\midrule
\textbf{0.01} &\textbf{0.1} &\textbf{1} &\textbf{1} &\textbf{5} &\textbf{66.8} &\textbf{0.201} \\
\bottomrule
\end{tabular}
\caption{The  best  top-3  R-precision score of each LAViTeR model setting on COCO (top nine rows) and CUB (bottom 6 rows) test set. AIMCoS score is also reported.}\label{tab: hyperparam}
\end{table}

As we can see, different $\lambda$ settings have influenced the performance of the model. The best setting on CUB is from the bottom row in Table \ref{tab: hyperparam}, which gives the best top-3 R-precision as 66.8\%, where we choose a relative large $\lambda_m=5$ for loss $L_m$ while keeping small weights as $\lambda_G=0.01$ and $\lambda_C=0.1$ for $L_G$ and $L_C$. Meanwhile, row 9 in Table \ref{tab: hyperparam} with a similar setting ($\lambda_m=10, \lambda_G=0.01, \lambda_C=0.1$) also gives the best top-3 R-precision on COCO as 90.79\%, and row 3 in Table \ref{tab: hyperparam} which adjusts $\lambda_m$ to 50 gives the best AIMCoS score as 0.5063. These results show that $L_m$ requires higher weight among all the loss terms, since it's a credible objective for real-image-real-text matching. On the other hand, $L_m^I$ and $L_m^T$ are calculated from the fake image or fake text matching, which may induce some noises in the gradients due to some low quality generated images and captions, thus smaller $\lambda_{\hat{I}}$ and $\lambda_{\hat{T}}$ gives better results. Moreover, since $L_G$ and $L_C$ is only used for necessary updates in $G$ and $C$ during joint training and has limited contribution to VTA, thus we just keep tiny weights for them. 

\textbf{Ablation Studies and Analysis} We also conduct necessary ablation studies in Table \ref{tab: coco_horizontal} to further inspect the effect of each modules of LAViTeR. First, we set all the $\lambda$s to 0 except $\lambda_m$, which results in the worst R-precision among all the settings on COCO dataset (see the top row in Table \ref{tab: hyperparam}). This case shows the importance of our assisting losses during the joint training. Since $L_m$ trains the image and text encoders but has no effect on $C$ and $G$, the updated feature outputs from $F_I$ and $F_T$ are no longer working well for the captioning model and image generator, hence the quality of generated images and captions degraded which also affects the matching performance. 

In Table \ref{tab: coco_horizontal}, we test the performance of LAViTeR when different modules are removed. Since symmetric global-local matching loss $L_m$ is also used to train the image-text matching model (called DAMSM in their work) in \cite{xu_attngan, qiao2019mirrorgan}, so we also evaluate their performance using our metrics, which refers to AttnGAN/MirrorGAN in Table \ref{tab: coco_horizontal} and \ref{tab: cub_horizontal}. DAMSM uses RNN-based text encoder in their paper instead of the transformer-based model like ours, so it also plays a role as one of our baseline. In our ablation settings, LAViTeR is our final model. $L_M(F)$ and $L_M(T)$ mean that only the VTA module is trained in phase 3 as our baseline without TIM and ITM branches, and the image encoder backbone is frozen ($F$, same training method as DAMSM) and trainable ($T$) during the joint training, respectively. LAViTeR-Img2Txt only keeps the ITM module in the joint training to generate image captions for $L_m^T$ and the image generator $G$ with $L_m^I$ is removed; As a complementary setting, LAViTeR-Txt2Img keeps the image generator and $L_m^I$ and deducts the captioning module $C$ along with $L_m^T$. 

As it shows in Table \ref{tab: coco_horizontal}, $L_M(T)$ improves the performance for 0.66\% in R-precision and 0.005 in AIMCoS, which shows that the image encoder requires update during training for a better performance. In LAViTeR-Img2Txt where $G$ is trimmed, the R-precision drops by 0.57\% from the full model and even lower than the baseline $F_M(T)$. This might be related to the large amount of unseen fake images no longer being generated and used for VTA training, thus reduces the diversity of the training set and the matching accuracy. Meanwhile, if $C$ is removed as in LAViTeR-Txt2Img setting, both the R-precision and AIMCoS are reduced by 0.28\% and 0.0023 respectively, which could be caused by lack of fake captions. When no captions are generated for matching, the training sentences and words become less various and thus affect the image-text matching performance shown in R-precision as well as the word level matching accuracy suggested by AIMCoS. On both CUB and COCO datasets, our full model LAViTeR works much better than DAMSM in \cite{xu_attngan, qiao2019mirrorgan}, which improves R-precision by 0.74\% in COCO and 4.2\% in CUB, and boosts AIMCoS from 0.44 to 0.5063 in COCO and 0.072 to 0.201 in CUB, almost 3 times better. This result shows that image-to-text and text-to-image generation branches can assist the image-text matching/representation model training by aligning generated image with real text and 
generated text with real image. 

We also find an interesting phenomenon in our baselines. When replacing the original RNN-based text encoder in DAMSM with our transformer-based text encoder, $L_M(F)$ has a slightly lower R-precision but much higher AIMCoS than DAMSM baseline in both Tables. This actually shows one of the main difference between transformer layer and RNN layer: due to the recurrence feature and short memory of RNN, one layer transformer may have no obvious advantage over RNN for captions with normal length around 15, which is implied in R-precision for image-caption matching; however, RNN cannot get good representations of short phrases with only 1-3 words, since it requires enough context to give reasonable outputs. In contrast, transformer's self-attention mechanism is able to effectively handle any text length and align even a single word with the corresponding image regions, which is observed in the substantial increase in AIMCoS. 

\begin{table}[!htp]\centering
\small
\begin{tabular}{lccc}\toprule
\textbf{Model Name} &\textbf{R-precision} &\textbf{AIMCoS} \\\midrule
AttnGAN/MirrorGAN\cite{xu_attngan,qiao2019mirrorgan} &89.97 &0.44 \\
LAViTeR-$L_{M}$ (F) &89.81 &0.501 \\
LAViTeR-$L_{M}$ (T) &90.37 &0.5059 \\
LAViTeR-Img2Txt &90.14 &0.506 \\
LAViTeR-Txt2Img &90.43 &0.504 \\
\midrule
\textbf{LAViTeR} &\textbf{90.71} &\textbf{0.5063} \\
\bottomrule
\end{tabular}
\caption{Top-3 R-precision and Attribute Image Match Cosine Score (AIMCoS) metrics calculated on COCO dataset.}\label{tab: coco_horizontal}
\end{table}

\begin{table}[!htp]\centering
\small
\begin{tabular}{lccc}\toprule
\textbf{Model Name} &\textbf{R-precision} &\textbf{AIMCoS} \\\midrule
AttnGAN/MirrorGAN \cite{xu_attngan,qiao2019mirrorgan} &62.6 &0.072 \\
LAViTeR-$L_M$ (F) &59.2 &0.184 \\\midrule
\textbf{LAViTeR} &\textbf{66.8} &\textbf{0.201} \\
\bottomrule
\end{tabular}
\caption{Top-3 R-precision and Attribute Image Match Cosine Score (AIMCoS) metrics calculated on CUB dataset.}\label{tab: cub_horizontal}
\end{table}

\begin{figure}[ht]
\begin{center} 
  \includegraphics[width=1.0\linewidth]{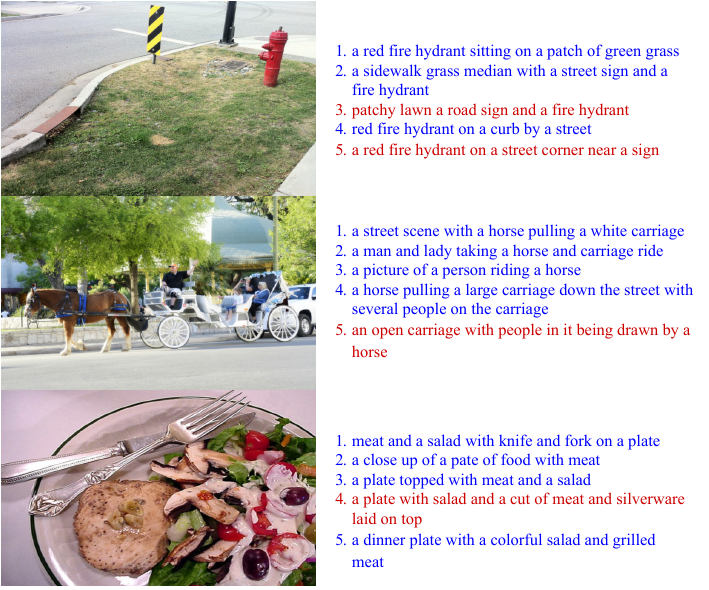}
\end{center}
\caption{The top-5 image-to-text matching captions with descending similarity scores. Blue captions are the correct matches, while red ones are incorrect matches.}
\label{fig:i2t_vis}
\end{figure}

\begin{figure}[ht]
\begin{center} 
  \includegraphics[width=1.0\linewidth]{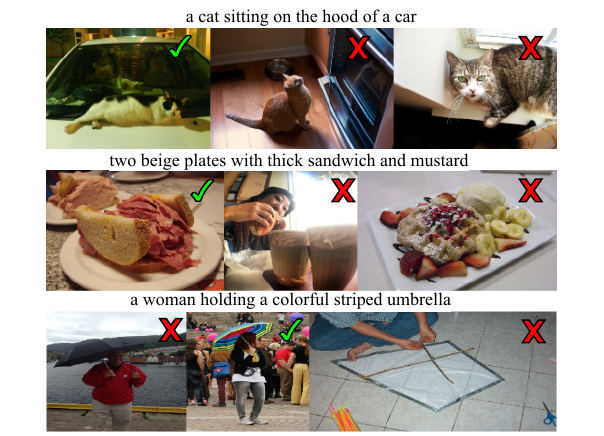}
\end{center}
\caption{The top-3 text-to-image matching images with descending similarity scores from left to right. Green marks are the correct matches, while red crosses are incorrect matches.}
\label{fig:t2i_vis}
\end{figure}

\textbf{Visualization Analysis}
In addition, we illustrate some qualitative results from image-to-text and text-to-image matching on COCO in Figure \ref{fig:i2t_vis} and Figure \ref{fig:t2i_vis} respectively. For image queries, the captions with top-5 similarity scores from our model are retrieved. Most captions are correct matches. It is observed that sentences that have ``incorrect'' match labels actually share similar semantics with the image queries. In Figure \ref{fig:t2i_vis}, text queries with the top-3 image matches from our model are listed. The correct matching is retrieved with other similar images, and we find that all the high ranked images are quite reasonable. These ``incorrect'' matching pairs with close semantics data expose a drawback in the current evaluation metrics for multi-modal matching: many images along with their paired captions in the test set share similar semantics and are close to each other in the common semantic space, however, these neighbor samples are treated as ``mismatch'' in the image-to-text and text-to-image retrieval evaluation metrics and show no difference with those distinctly mismatched samples with large semantic gap. This evaluation defect motivates us to propose the new evaluation metric AIMCoS.

\section{Conclusion}
In this paper, we introduce a novel architecture for visual and textual representation learning assisted by two auxiliary tasks, image-to-text generation and text-to-image generation. The generated images and texts are matched with real text and images to jointly train the representation model with two assisting matching losses. A new evaluation metric AIMCoS is proposed for measuring the similarity between the learnt visual and textual embedding. The experimental results on two public datasets demonstrate the effectiveness of the proposed architecture and evaluation metric. 

{\small
\bibliographystyle{ieee}
\bibliography{references}
}

\newpage

\appendixpage
\appendix
Below we mention some analysis and downstream tasks performed using our pre-trained encoders that we include as supplementary materials.

\section{VTA Qualitative Results}

\begin{figure}[ht]
\begin{center}
\includegraphics[width=1.0\linewidth]{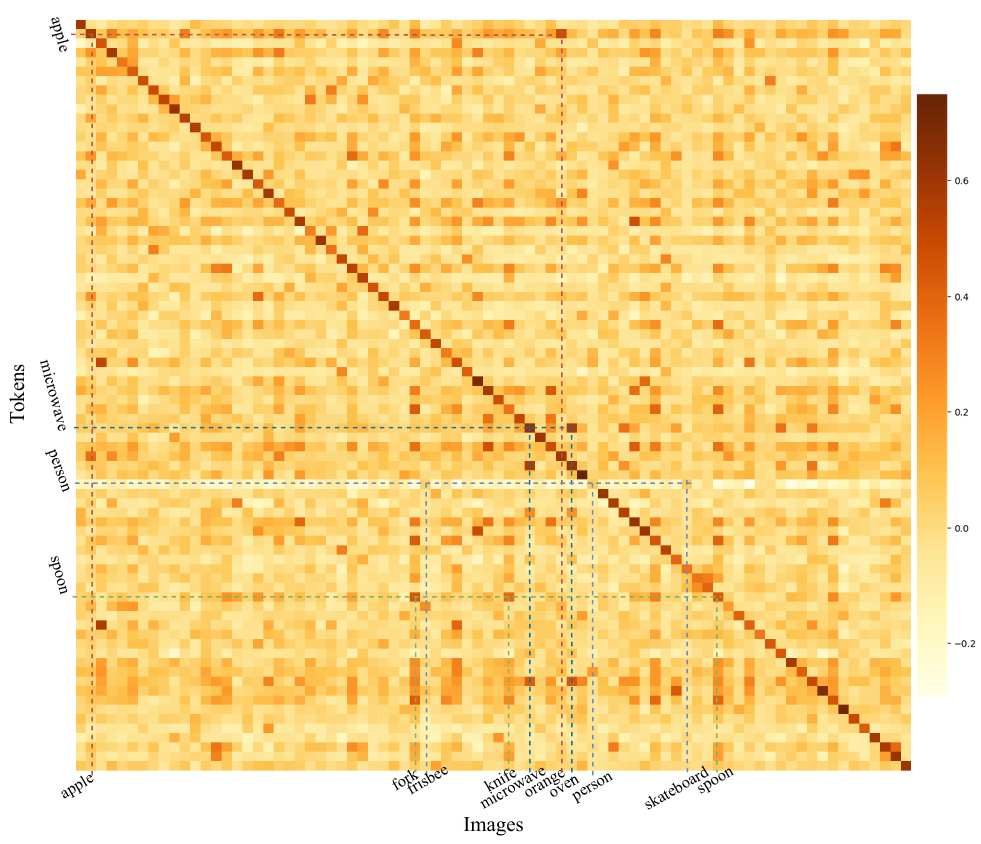}
\end{center}
   \caption{Visualization of Image representation vs Token representation similarity map}
\label{fig:similarity_map}
\end{figure}

Apart from the T-SNE visualization in main paper, we also plot the Image representation vs Token representation similarity map in Figure \ref{fig:similarity_map}. 

The x-axis denotes the image representations and the y-axis denotes the category word token representation. Since there are 80 classes in COCO dataset \cite{coco2014}, the matrix size is $80 \times 80$. There are 80 Tokens and 8000 images, i.e. 100 images per token. We first compute similarity of each token representation with all the 8000 images in LaViTeR\textsubscript{cocoeval} dataset. Then, the average similarity value of match with each token plotted in Figure \ref{fig:similarity_map}. The dark color diagonal shows a high similarity between true matches, indicating the efficacy of LAViTeR model. 

From Figure \ref{fig:similarity_map}, we observe that, the token ``apple'' matches highest with the images of apple but also contains high similarity with images of another fruits like ``orange'' because they are usually kept together. Similarly, the token ``microwave'' is highly matched with images of ``microwave'' and also with ``oven'' and in reality most of the times they appear together. More interestingly, the token ``person'' is not matched with other images showing no correlation, but is matched highly with images of ``person'' and also ``skateboard'' as in many ``skateboard'' images there is a person in the frame. Also, the token ``spoon'' is highest matched with ``spoon'' images and also slightly matched with images of ``fork'' and ``knife'' as these items usually appear together naturally.
Thus, the joint training is able to reduce the diversity gap between the nature of textual and visual representations meaningfully.

\section{TIM Qualitative Results}
Although our primary goal is not to improve GAN, we compare the quality of generated images from LAViTeR model with the ground truth and the generated images by \cite{qiao2019mirrorgan,xu_attngan}. Those examples are selected from the visualization of the previous state-of-the-art papers directly. 
\begin{figure}[ht]
\begin{center}
\includegraphics[width=1.0\linewidth]{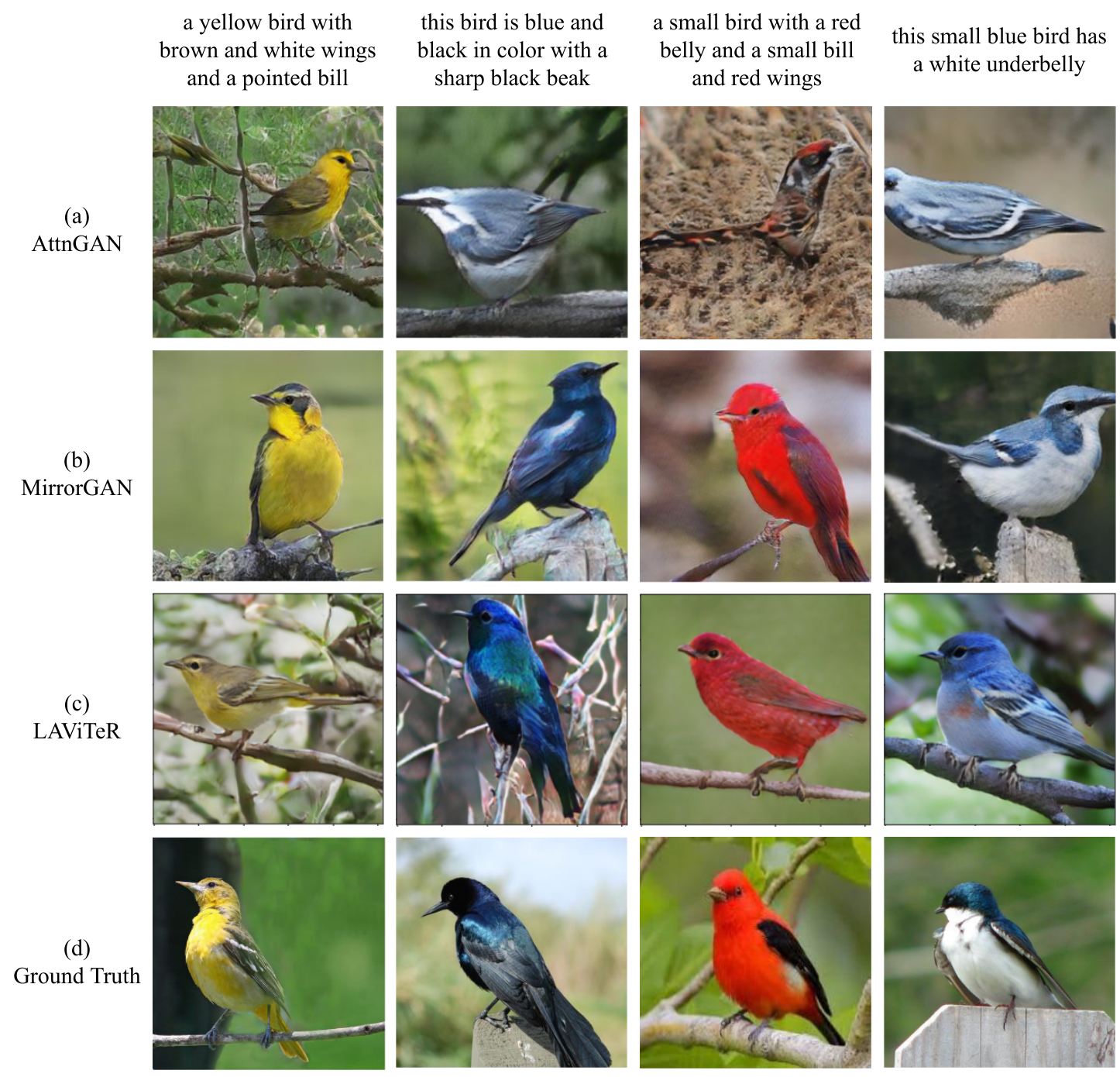}
\end{center}
   \caption{Examples of images generated by (a) AttnGAN \cite{xu_attngan}, (b) MirrorGAN \cite{qiao2019mirrorgan} (c) LaViTeR network (d) the corresponding ground truth. Left four columns are images from CUB \cite{cub2011} dataset.}
\label{fig:qualtiy_cub}
\end{figure}

\begin{figure}[ht]
\begin{center}
\includegraphics[width=1.0\linewidth]{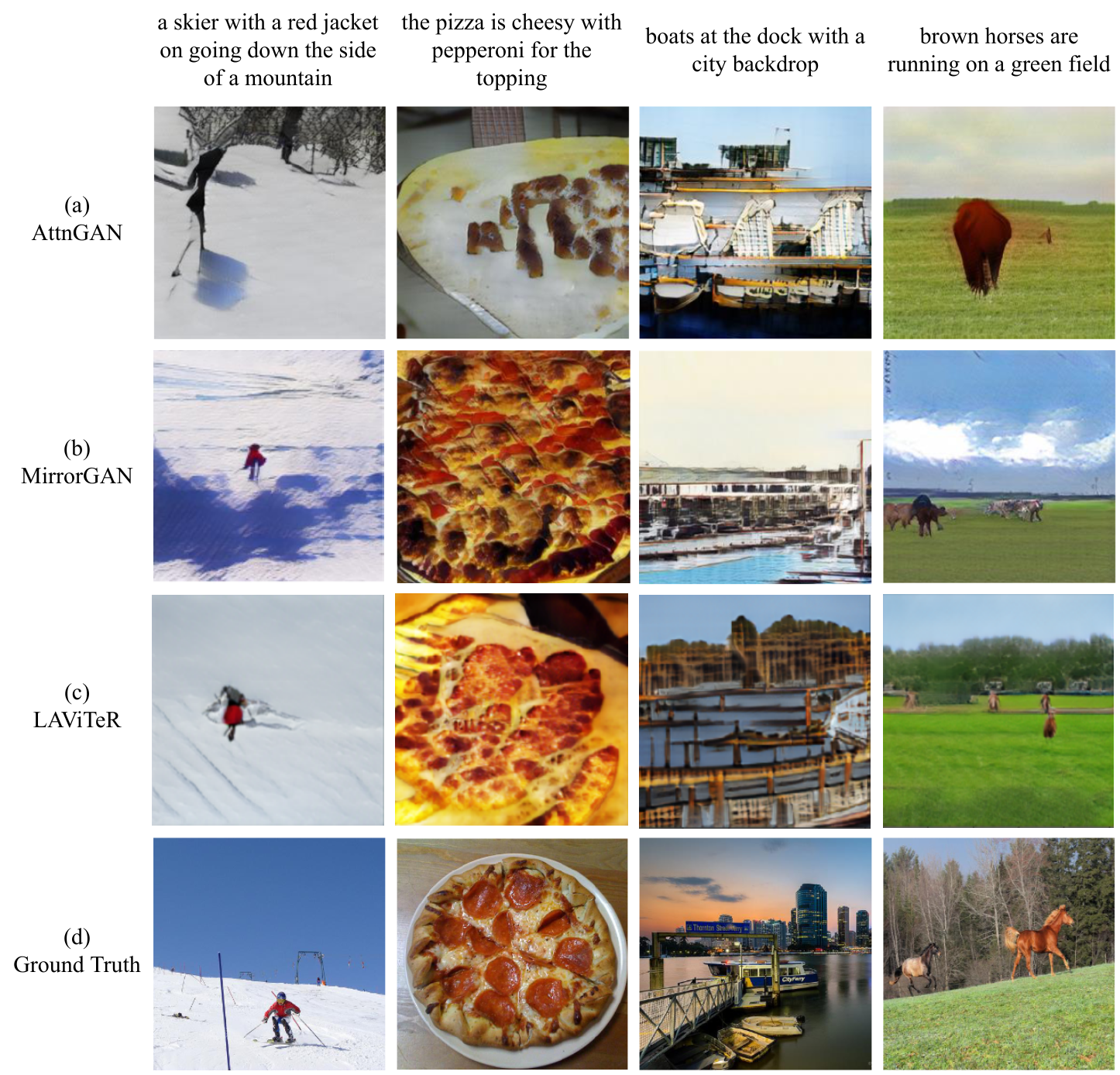}
\end{center}
   \caption{Examples of images generated by (a) AttnGAN \cite{xu_attngan}, (b) MirrorGAN \cite{qiao2019mirrorgan} (c) LaViTeR network (d) the corresponding ground truth. Left four columns are images from COCO \cite{coco2014} dataset.}
\label{fig:qualtiy_coco}
\end{figure}

As can be seen from Figure \ref{fig:qualtiy_cub} and \ref{fig:qualtiy_coco}, the LAViTeR model is able to perform at-par and in some cases better than the previous state-of-the-art models. We do not report these images in the main paper as the major goal of LaViTeR is for representation joint learning and alignment. The model is able to perform better due to continuous training of VTA, TIM and ITM models cooperatively. The generated images during the training process are able to provide a diverse set of images, which further assists the joint learning and also the GAN module.

\begin{table}[ht]\centering

\begin{tabular}{lrrr}\toprule
\textbf{Model} &\textbf{IS} $\uparrow$ &\textbf{FID} $\downarrow$ \\\midrule
StackGAN-v1 &8.45 $\pm$ 0.03 &74.05 \\
StackGAN-v2 &8.30 $\pm$ 0.10 &81.59 \\
AttnGAN &25.89 $\pm$ 0.47 &- \\
MirrorGAN &26.47 $\pm$ 0.41 &- \\\midrule
LAViTeR &26.71 $\pm$ 0.39 &75.5 \\
\bottomrule
\end{tabular}
\caption{IS and FID scores, calculated on COCO 2014 validation set. Uparrow means higher value is better, down arrow means lower value better.}\label{tab:gan_scores}
\end{table}
We also compute the Inception Score (IS) \cite{inceptionscore} and Fr\'echet Inception Distance (FID) \cite{fid} to compare the GAN module trained while joint learning. Table \ref{tab:gan_scores} shows that GAN in LaViTeR model performs better than all the previous  models in terms of IS Score while the FID score is still comparable to other text to image network as \cite{stackGan} that report FID scores. However, better GAN performance is a by-product of LAViTeR model, not it's goal.

\section{ITM Qualitative Results}

\begin{figure}[h!]
\begin{center}
\includegraphics[width=1.0\linewidth]{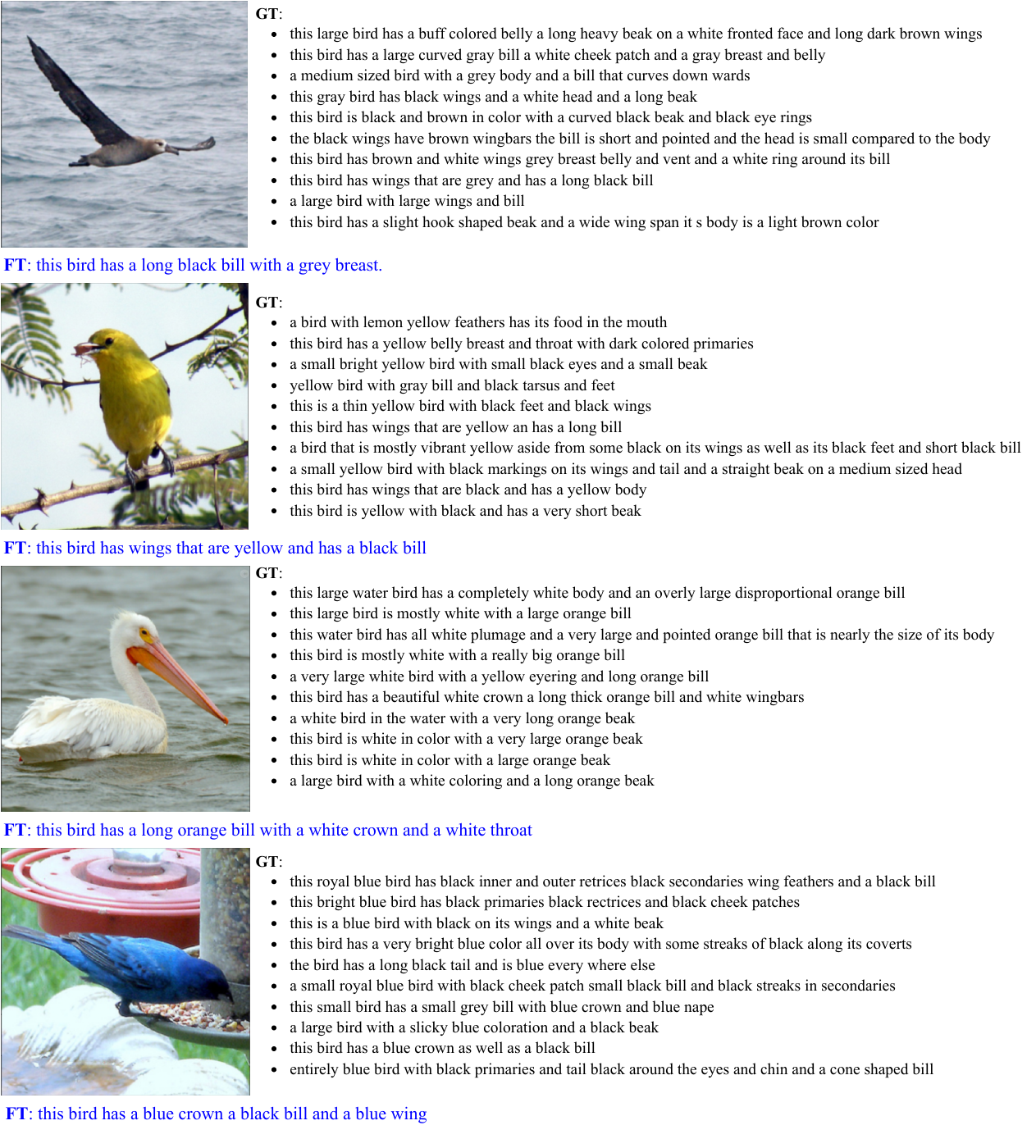}
\end{center}
   \caption{Examples of generated captions (FT (Fake Text) in blue under sample images) by LAViTeR ITM and the corresponding ground truth captions (GT in black).}
\label{fig:cap_cub}
\end{figure}

\begin{figure*}[ht!]
\begin{center}
  \includegraphics[width=1.0\linewidth]{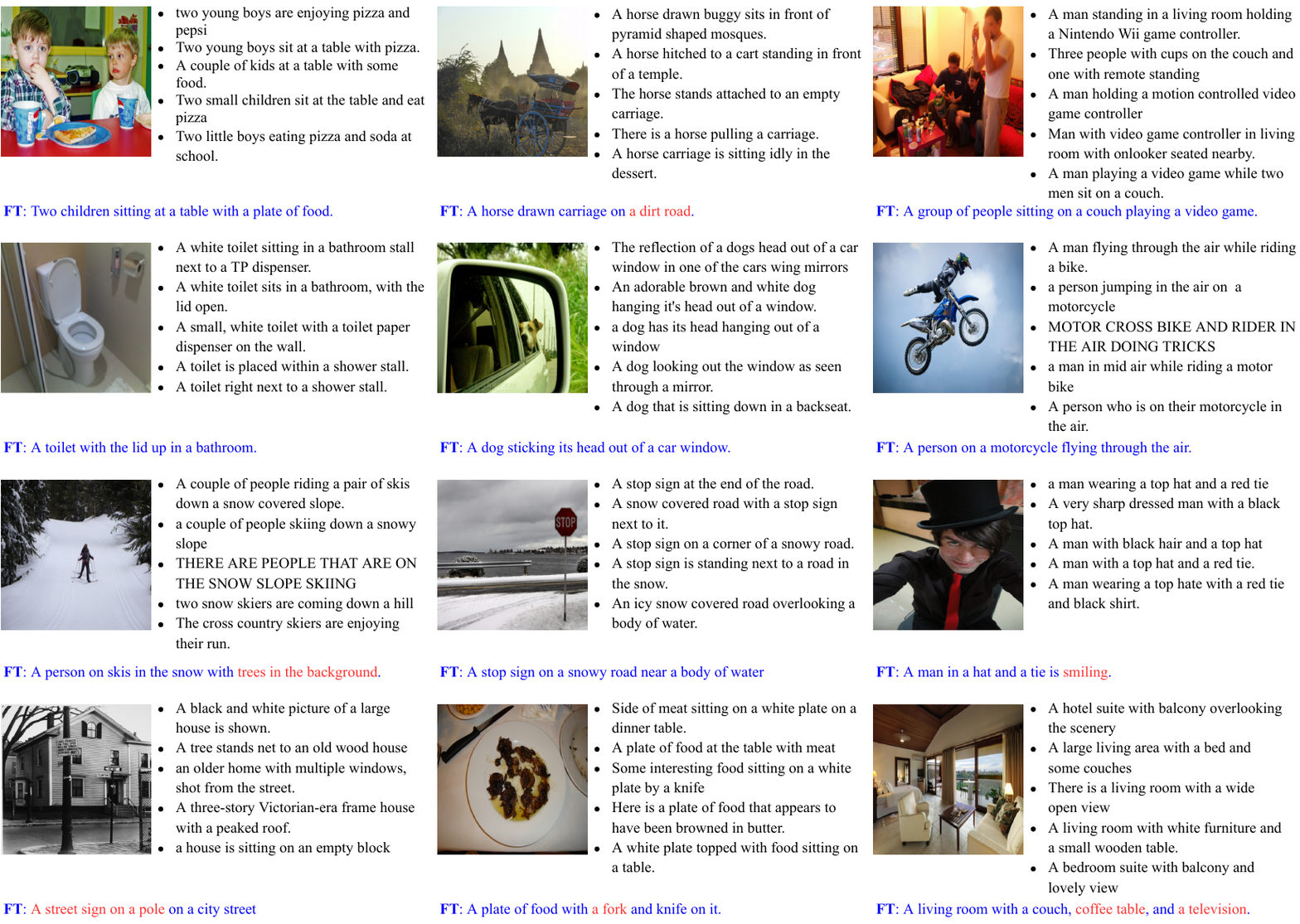}
\end{center}
  \caption{Examples of generated captions (FT (Fake Text) in blue under sample images) by LaViTeR ITM and the corresponding ground truth captions (in black). The objects or attributes shown in the sample image but not in its ground truth descriptions are marked in red.}
\label{fig:cap_coco}
\end{figure*}

\begin{table*}[h!]\centering
\small
\begin{tabular}{lrrrrrr}\toprule
\textbf{Model} &\textbf{BLEU-1} &\textbf{BLEU-2} &\textbf{BLEU-3} &\textbf{BLEU-4} &\textbf{METEOR} &\textbf{CIDEr}\\\midrule
Show-tell \cite{vinyals2015show} &- &- &- &27.7 &23.7 &85.5\\
Show-attend-tell \cite{xu2015show} &70.7 &49.2 &34.4 &24.3 &23.9 &- \\\midrule
LAViTeR &69.15 &51.75 &37.68 &27.44 &24.20 &86.72 \\
\bottomrule
\end{tabular}
\caption{BLEU 1-4, METEOR\cite{denkowski2014meteor} and CIDAr\cite{vedantam2015cider} scores of our ITM module calculated on MSCOCO 2014 validation set. Results reported in \cite{vinyals2015show} and \cite{xu2015show} are also listed here for reference.}\label{tab:cap_scores}
\end{table*}

Some examples of generated captions by our LAViTeR ITM branch are shown in Figure \ref{fig:cap_cub} and Figure \ref{fig:cap_coco}. As Figure \ref{fig:cap_cub} shows, the generated captions are able to capture the main attributes (crown color, bill size, breast color, etc.) of the birds in sample images from CUB test set. Figure \ref{fig:cap_coco} shows objects and their corresponding actions and attributes in sample images from COCO test set. Specifically, some of the generated captions can generate words (marked in red) that are not included in the ground truth captions, but actually shown up in the images. Some objects in the background are predicted in the generated captions, e.g. for the first image in the third row, `trees' is in the background of the image which is never described in the paired ground truth captions, but successfully predicted by ITM; for the last image, TV and coffee table are shown in the background corner of the image and are ignored by the ground truth texts, but are captured by the generated caption. Even only a part of the object appears in an image, it still has a chance to show up in our generated caption, e.g. in the image in the middle of the last row, a tiny part of a fork tip appears on the left, our ITM can generate `a fork' given such a small clue in the image. Missing attributes are also predicted such as the attribute `smiling' is generated in our generated text, which is never described in the ground truths. These generated texts from ITM with the missing objects and attributes in the original images provide further useful information outside the training dataset and help the matching model learn more word-region level matching pairs. This is exactly one of our motivations for our LAViTeR method. 

We also evaluate the captioning ability of our ITM branch from the joint training, listed in Table \ref{tab:cap_scores}. As we can see, our ITM can get a comparable result with other captioning models, such as \cite{xu2015show} and \cite{vinyals2015show}, which shows that our ITM has the ability to generate reasonable captions for real-image-fake-text matching training. Note that we are not aiming at coming up with a novel and state-of-the-art captioning model, the main function of ITM branch is to provide more reasonable fake text for training the VTA matching model. 

\section{Code}
Our code referenced from libraries open sourced by the work done in CATR \cite{catr} and AttnGAN \cite{xu_attngan}.

\end{document}